\documentclass[10pt,twocolumn,letterpaper]{article}

\usepackage{cvpr}
\usepackage{times}
\usepackage{epstopdf}
\usepackage{graphicx}
\usepackage{amsmath}
\usepackage{amsbsy}
\usepackage{amssymb}
\usepackage{times}
\usepackage{helvet}
\usepackage{courier}
\usepackage{amsmath}                                    
\usepackage[final]{pdfpages}
\usepackage{amssymb}
\usepackage[linesnumbered, ruled, boxed]{algorithm2e}
\newtheorem{theorem}{Define}
\usepackage{multirow}
\newtheorem{myTheorem}{Theorem}
\usepackage{pifont}
\cvprfinalcopy

\usepackage[pagebackref=true,breaklinks=true,letterpaper=true,colorlinks,bookmarks=false]{hyperref}

\def\cvprPaperID{974} 

\newcommand{\nosection}[1]{\vspace{3pt}\noindent\textbf{#1.}}
\ifcvprfinal\fi
\begin{document}

\title{P3SGD: Patient Privacy Preserving SGD \\ for Regularizing Deep CNNs in Pathological Image Classification }

\author{{Bingzhe Wu$^{1}$\thanks{Work performed while interning at IBM Research - China.}}\qquad
		{Shiwan Zhao$^{2}$}\qquad
		{Guangyu Sun$^{1}$}\qquad
		{Xiaolu Zhang$^{3}$}\\
		{Zhong Su$^{2}$}\qquad
		{Caihong Zeng$^{4}$}\qquad
		{Zhihong Liu$^{4}$}
		\vspace{2mm}\\
		{$^1$Peking University}\; {$^2$IBM Research}\; {$^3$Ant Financial Service Group}\\
		{\small \tt \{wubingzhe,gsun\}@pku.edu.cn},\; {\small \tt \{zhaosw,suzhong\}@cn.ibm.com},\; {\small \tt yueyin.zxl@antfin.com}\\
		{$^4$National Clinical Research Center of Kidney Diseases, Jinling Hospital}\\
		{\small \tt zengch\_nj@hotmail.com},\; {\small \tt liuzhihong@nju.edu.cn}
	}
\maketitle
\thispagestyle{empty}

\begin{abstract}
\vspace{-0.3cm}
Recently, deep convolutional neural networks~(CNNs) have achieved great success in pathological image classification. However, due to the limited number of labeled pathological images, there are still two challenges to be addressed: (1) overfitting: the performance of a CNN model is undermined by the overfitting due to its huge amounts of parameters and the insufficiency of labeled training data. (2) privacy leakage: the model trained using a conventional method may involuntarily reveal the private information of the patients in the training dataset. The smaller the dataset, the worse the privacy leakage.

To tackle the above two challenges, we introduce a novel stochastic gradient descent~(SGD) scheme, named patient privacy preserving SGD~(P3SGD), which performs the model update of the SGD in the patient level via a large-step update built upon each patient's data. Specifically, to protect privacy and regularize the CNN model, we propose to inject the well-designed noise into the updates. Moreover, we equip our P3SGD with an elaborated strategy to adaptively control the scale of the injected noise. To validate the effectiveness of P3SGD, we perform extensive experiments on a real-world clinical dataset and quantitatively demonstrate the superior ability of P3SGD in reducing the risk of overfitting. We also provide a rigorous analysis of the privacy cost under differential privacy. Additionally, we find that the models trained with P3SGD are resistant to the model-inversion attack compared with those trained using non-private SGD.

\end{abstract}
\section{Introduction}
\begin{figure}
    \centering 
    \includegraphics[width=7cm]{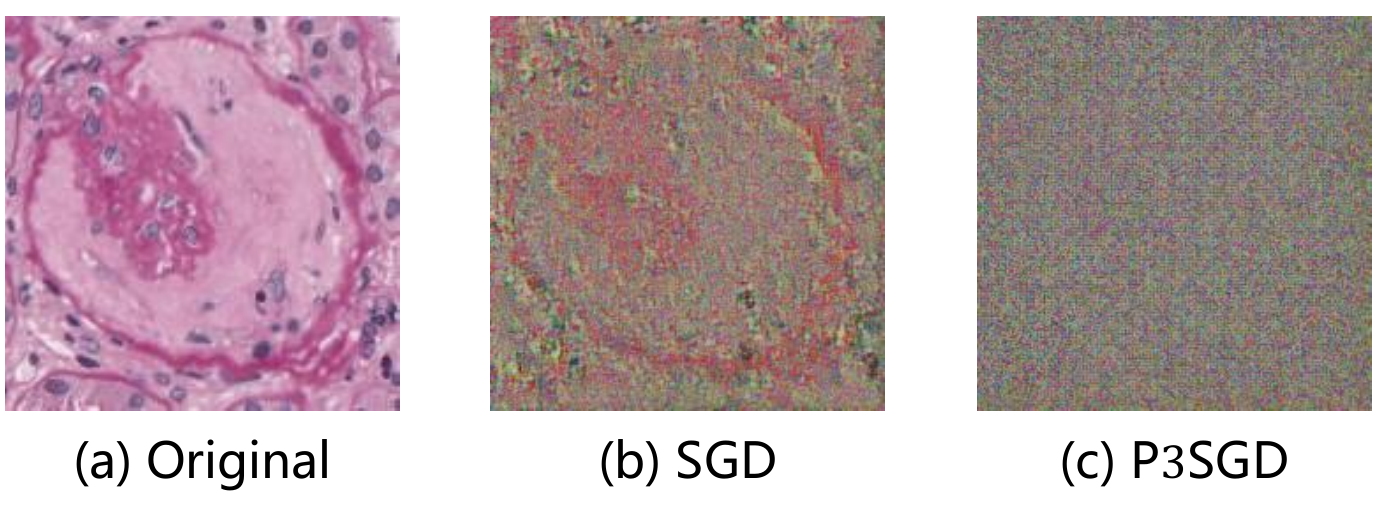}
    \caption{Illustration of the model-inversion attack on ResNet-18. (a) is the original image patch. (b) and (c) are the images reconstructed from the model trained with non-private SGD and P3SGD.}
    \label{fig:model_inversion1}
    \vspace{-0.5cm}
\end{figure}

In recent years, deep CNNs have emerged as powerful tools for various pathological image analysis tasks, such as tissue classification~\cite{pedraza2017glomerulus, Gallego2018Glomerulus}, lesion detection~\cite{Janowczyk2016Deep, liu2017detecting}, nuclei segmentation~\cite{Nuclei_seg_2016, Nuclei_seg_2015, Nuclei_seg_2017}, etc. The superior performance of deep CNNs usually relies on large amounts of labeled training data~\cite{2017dnnsurvey}. Unfortunately, the lack of labeled pathological images for some tasks may lead to two notorious issues: (1) overfitting of the CNN models~\cite{dropout14, Dropconnect, Zhang2016a} and (2) privacy leakage~\cite{Zhang2016a, memorization, Fredrikson2015, csf18_relation} of the patients. Firstly, the performance of a CNN-based model is always harmed by the overfitting due to its large amounts of parameters and the insufficiency of training data. Secondly, pathological datasets usually contain sensitive information, which can be associated with each individual patient. The CNN-based models trained using conventional SGD  may involuntarily reveal the private information of patients according to recent studies~\cite{Zhang2016a,Fredrikson2015}. For example, Zhang~\etal \cite{Zhang2016a} show that the CNN model can easily memorize some samples in the training dataset. Fredrikson~\etal \cite{Fredrikson2015} propose a model-inversion attack to reconstruct images in the training dataset.
In Figure~\ref{fig:model_inversion1} (a) and (b), we demonstrate an attacking example in our task, reconstructing the outline of a patch in the training dataset by leveraging a well-trained CNN model and its intermediate feature representations.

There have been numerous studies to solve either of two issues individually. On the one hand, to reduce the risk of overfitting in deep CNNs, previous research suggests adding appropriate randomness into the training phase~\cite{dropout14, Dropconnect, xie2016disturblabel}. For example, Dropout~\cite{dropout14} adds randomness in activation by randomly discarding the hidden layers' outputs. DropConnect~\cite{Dropconnect} adds randomness in weight parameters by randomly setting weights to zero during training. On the other hand, differential privacy~\cite{dwork_dp_2006, DP_2008} emerges as a strong standard, which offers rigorous privacy guarantees for algorithms applied on the sensitive database. Recent works~\cite{abadi2016deep, private_deep_vae} are
introduced to train deep CNN models within differential privacy. The main idea of these works is to perturb the gradient estimation at each step of an SGD algorithm. For example, Abadi et al.~\cite{abadi2016deep} use a differentially private additive-noise mechanism on the gradient estimation in an SGD.
In addition, a few recent studies~\cite{memorization, csf18_relation} have shown that these two seemingly unrelated
issues are implicitly relevant based on a natural intuition: \emph{``reducing the overfitting''} and \emph{``protecting the individual's privacy''} share the same goal of encouraging a CNN model to learn the population's features instead of memorizing the features of each individual. 

In this paper, we propose a practical solution to alleviate both issues in a task of pathological image classification. In particular, we introduce a novel SGD algorithm, named P3SGD, which injects the well-designed noise into the gradient to obtain a degree of differential privacy and reduce overfitting at the same time. It is worth noting that a pathological database usually consists of a number of patients, each of whom is further associated with a number of image patches. We should protect the privacy in the \textbf{patient level} instead of image level as in most of the previous works. To achieve this goal, we propose to calculate the model update upon individual patient's data and add carefully-calibrated Gaussian noise to the update for both privacy protection and model regularization. 
The most similar work to ours is the differentially private federated learning~\cite{brendan2018learning,geyer2017differentially}, which
focuses on protecting the user-level privacy. 
In contrast to previous works, which use a globally fixed noise scale to build the noisy update~\cite{abadi2016deep, brendan2018learning, geyer2017differentially}, we propose an elaborated strategy to adaptively control the magnitude of the noisy update. In the experiment, we show that this strategy plays a key role in boosting performance of a deep CNN model. At last, we provide a rigorous privacy cost analysis using the moments accountant theorem~\cite{abadi2016deep}. 



In summary, the main contributions of our work are as follows:
\begin{itemize}
\vspace{-0.2cm}
\item We introduce a practical solution, named P3SGD, to simultaneously address the overfitting and privacy leaking issues of deep CNNs in the pathological image classification. To the best of our knowledge, this is the first work to provide rigorous privacy guarantees in medical image analysis tasks.
\vspace{-0.2cm}
\item Technically, we present a strategy to dynamically control the noisy
update at each iterative step, which leads to a significant performance gain against the state-of-the-art methods~\cite{brendan2018learning, geyer2017differentially}.
\vspace{-0.2cm}
\item We validate P3SGD on a real-world clinical dataset, which is less explored
in previous studies. The results demonstrate that P3SGD is capable of reducing the risk of overfitting on various CNN architectures. Moreover, 
P3SGD provides a strong guarantee that the trained model protects the
privacy of each patient's data, even when the attacker holds enough extra side-information of the raw training dataset.
\vspace{-0.2cm}
\item We qualitatively and quantitatively demonstrate that the CNN model trained
using P3SGD is resistant to the model-inversion attack~\cite{Fredrikson2015} (shown in Figure~\ref{fig:model_inversion1} (c)).
\end{itemize}

\section{Related Work}
\label{section:related_work}
\noindent \textbf{Regularization in CNNs}$\quad$In the past years, numerous regularization techniques have been proposed to improve the generalization ability of deep CNNs~\cite{2009weight_decay, dropout14, Dropconnect, xie2016disturblabel, 2015SpatialDropout, 2018dropblock}.
These works mainly fall into two categories: explicit regularization and implicit~(\emph{i.e.}, algorithmic) regularization.

For explicit regularization methods, various penalty terms are used to constrain weight parameters. For example, weight decay~\cite{2009weight_decay}  
uses $l2$-regularization to constrain the parameters of a CNN model. Another direction is to introduce regularizers to decorrelate convolutional filters in deep CNNs~\cite{wu_decorrelate, orgreg17}, which improves
the representation ability of the intermediate features extracted by those filters.

For implicit regularization methods, the core idea is to introduce moderate randomness in the model training phase. For example, Dropout~\cite{dropout14} randomly discards the outputs of the hidden neurons in the training phase. 
However, Dropout is originally designed for fully-connected layers~(FC). It is often less effective for convolutional layers, which limits 
its use in CNNs with few FC layers (\emph{e.g.}, ResNet). 
This is possibly caused by the fact that Dropout discards features without taking its spatial correlation into account (features from convolutional layers are always spatially correlated)~
\cite{2018dropblock}. To address this problem, a few recent works~\cite{2015SpatialDropout, 2018dropblock} propose to inject structured noise into the features from convolutional layers. One state-of-the-art 
technique, named DropBlock~\cite{2018dropblock}, is specially designed for convolutional layers, which randomly drops the features in a sub-region. 
Both of Dropout and DropBlock inject randomness into activation layers. 
In contrast, DisturbLabel~\cite{xie2016disturblabel} adds randomness into the loss function by randomly setting a part of labels to be incorrect in a training mini-batch. 
Data augmentation is another form of algorithmic regularization, which introduces noise into the input layer by randomly transforming training images~\cite{Simonyan15}. 
Our method can be categorized as implicit regularization. In contrast to previous works, our approach (P3SGD) imposes regularization at the parameter updating phase.

\noindent \textbf{Privacy-preserving Deep Learning}$\quad$Meanwhile, there is an increasing concern for privacy leakage in deep learning models, since the training datasets may contain sensitive information. This privacy issue has attracted many research interests
on the privacy-preserving deep learning~\cite{abadi2016deep,shokri2015privacy,brendan2018learning, geyer2017differentially, 2016cryptonets, pate}. One promising direction is to build  machine learning models within differential privacy~\cite{abadi2016deep,brendan2018learning, pate}, which
has been widely used in sensitive data analysis as a golden standard of privacy. The early solution is to perturb the model
parameters~\cite{private_lr09,private_obj_zhang} or the objective function~\cite{private_lr09, private_obj_colt12, private_deep_vae}. However, such kind of simple solutions cause considerable performance decreasing~\cite{private_erm, preivate_erm_nips17}, the situation may become worse in the context of 
deep learning. Therefore, some recent studies focus on the gradient perturbation based methods~\cite{abadi2016deep,brendan2018learning,geyer2017differentially, pate, scalable_pate}. Abadi~\etal\cite{abadi2016deep} propose a differentially private
version of SGD and present the moments accountant framework to provide tighter privacy bound than previous methods. The PATE framework~\cite{pate, scalable_pate} protects the privacy via transferring 
knowledge to the student model, from an ensemble of teacher models, which are trained on partitions of the training data.

Different from these works, which focus on image-level privacy, we aim to provide \textbf{patient-level} privacy in specific scenarios of pathological image analysis. The most similar works to ours are ~\cite{brendan2018learning, geyer2017differentially}, which
extend the private SGD into the federated learning paradigm~\cite{federated_learning}. However, applying these approaches to the real-world medical image data remains less explored. 
Moreover, these methods always lead to a performance drop compared with the models trained using non-private SGD. In this paper, we evaluate our method on a real-world pathological image dataset and show that the performance drop can be addressed by carefully controlling the noisy update using our strategy. 

There are also some studies aiming to explore the relationship between the overfitting and the privacy leakage issues from the perspective
of memorization~\cite{memorization,csf18_relation}. In this paper, we present a practical solution to alleviate these two related issues simultaneously.

\section{Our Approach}
\label{section:method}
In this section, we describe our approach in details and provide a rigorous privacy cost analysis using the moments accountant theorem~\cite{abadi2016deep}.

\subsection{Preliminaries}
We firstly introduce some 
basic notations and definitions of differential privacy corresponding to our specific task.

In our setting, the pathological image dataset can be regarded as a database $\mathcal{D}$ with $N_p$ patients.
Generally speaking, each patient $D_i$ consists of a number of image patches of various tissues, \emph{i.e.}, $\mathcal{D}_i = \{(\mathbf{x}_k, \mathbf{y}_k)\}^{N_i}_{k=1}$, where $N_i$ is the number of image patches of the $i$-th patient. With a slight abuse of notations, we also denote $\mathcal{D} =\bigcup_{i=1}^{N_p}\mathcal{D}_i$ as the whole set of images of all patients. Then, a basic concept of \textbf{image-level} adjacent databases can be defined as: two databases are adjacent if they differ in a single image-label pair~\cite{abadi2016deep}. This concept is widely used for image-level privacy protection.

However, such image-level privacy protection is insufficient for our tasks.   Instead, we introduce a concept of \textbf{patient-level} adjacent databases defined as follows:
\vspace{-0.2cm}
\begin{theorem}
\textbf{(Patient-level adjacent databases)}
 $\mathcal{D}'$ and $\mathcal{D}''$ are adjacent: if $\mathcal{D}'$ can be obtained
 by adding all images of a single patient to $\mathcal{D}''$ or removing all images of a single patient from $\mathcal{D}''$.
 \label{definition:adj}
\end{theorem}
This definition is inspired by the prior works~\cite{brendan2018learning, geyer2017differentially}, in which the authors focus on user-level privacy. With the definition of adjacent databases, we can formally define the patient-level differential privacy as:
\begin{theorem}
\label{definition:dp}
\textbf{(Differential privacy)} A randomized algorithm $\mathcal{A}: D\to~R$ satisfies
$(\epsilon, \delta)$-differential privacy if for any two adjacent databases $\mathcal{D}{'}, \mathcal{D}^{''}\subseteq D$ and for any subset of outputs $S \subseteq R$ it holds:

\begin{equation}
    \mathbf{Pr}[\mathcal{A}(\mathcal{D}^{'}) \in S] \leq e^{\epsilon} \mathbf{Pr}[\mathcal{A}(\mathcal{D}^{''}) \in S] + \delta
    \label{eq:dp}
\end{equation}

\end{theorem}
The randomized algorithm~$\mathcal{A}$ is
also known as the mechanism in the literature~\cite{dwork_dp_2006}. 
In our setting, $\mathcal{A}$ is the algorithm used to train deep CNNs, \emph{e.g.}, the SGD algorithm. $D$ denotes the training dataset (\emph{i.e.}, $\mathcal{D}$ in our case) and $R$ is the parameter space of a deep CNN. Intuitively, the Equation~\ref{eq:dp}
indicates that participation of one individual patient in a training phase has a negligible effect on the final weight parameters. 
 Another concept is the sensitivity of a randomized algorithm:
\begin{theorem}
\textbf{(Sensitivity)} The sensitivity of a randomized algorithm~$\mathcal{A}$ is the upper-bound
of $||\mathcal{A}(\mathcal{D}^{'})-\mathcal{A}(\mathcal{D}^{''})||_2$, where $\mathcal{D}^{'}$ and $\mathcal{D}^{''}$ are any adjacent databases~(see in Define~\ref{definition:adj}).
\label{definition:sensitivity}
\end{theorem}
To establish a randomized algorithm that satisfies differential privacy, we need to bound its sensitivity. 
The most used strategy is to clip the norm of the parameter update. In next two subsections, we will introduce the traditional SGD and P3SGD separately, as two instances of the randomized algorithm $\mathcal{A}$.

\subsection{Standard SGD Algorithm}
We start with the standard SGD (\emph{i.e.}, non-private SGD) algorithm for training a deep CNN-based classification model. The goal of the classification is to train a CNN
model $\mathbb{M}: {\hat{y}} = \mathbf{f(\mathbf{x}; \boldsymbol{\theta})}$, where $\hat{y}$ is the predicted label, and $\boldsymbol{\theta}$ are the model parameters. Training of the model is to minimize the empirical loss $\mathcal{L}(\mathcal{D};\boldsymbol{\theta})$. In practice, we estimate
the gradient of the empirical loss on a mini-batch. 
We denote the classification loss over a mini-batch as: 
\begin{equation}
\mathcal{L}(\mathcal{B}_t;\boldsymbol{\theta}) =  \dfrac{1}{|\mathcal{B}_t|}\sum_{(\mathbf{x},\mathbf{y}) \in \mathcal{B}_t} l(\mathbf{f}(\mathbf{x};\boldsymbol{\theta}), \mathbf{y})
\label{eq:sgd}
\end{equation}
Here, $l(\mathbf{x}, \mathbf{y})$ is the loss function, \emph{e.g.}, cross-entropy loss. $\mathcal{B}_t$ refers to a mini-batch of images which are randomly and independently drawn from the whole image set $\mathcal{D}$. Note that we can add an additional regularization term into Equation \ref{eq:sgd}, such as $l2$ term. 
At the $t$-th step of the SGD algorithm, we can update the current parameter $\boldsymbol{\theta}_t$ as 
 $\boldsymbol{\theta}_{t+1} = \boldsymbol{\theta}_t - \gamma_t \cdot \nabla_{\boldsymbol{\theta}_t}\mathcal{L}(\mathcal{B}_t;\boldsymbol{\theta}_t)$.

\subsection{P3SGD Algorithm}
Overall, our framework comprises of three components, which are \emph{update computation}, \emph{update sanitization}, and \emph{privacy accumulation}. 
Our method inherits the computing paradigm of federated learning~\cite{federated_learning}. Moreover, to protect the
privacy, we need to inject well-designed Gaussian noise into each step's
update, which is marked as \emph{update sanitization}. At last, 
we can use the moments accountant for \emph{privacy accumulation}.
The pseudo-code is depicted in Algorithm~\ref{algorithm:main}. Next, we will describe each of these components in details. 
\begin{algorithm}
\DontPrintSemicolon
\caption{P3SGD}
\label{algorithm:main}
\textbf{Inputs:} \\
\quad Patient database: $\mathcal{D}$, Empirical Loss: $\mathcal{L}$.\\
\quad Patient sampling ratio: $p$.\\
\quad Noise scale set $\Omega_{z}$: $\{z_i\}_{i=1}^{N_z}$.\\ 
\quad Noise budget $\epsilon'$ for selecting update per iteration.\\
\quad Bound of update's norm: ${\rm C_u}$.\\
\quad Bound of objective function's norm: ${\rm C_o}$.\\
\quad Initialize $\boldsymbol{\theta}_0$ randomly\\
\For{$t\in[T]$ }
{

Take a subset $\mathcal{B}_t$ of patients with sampling ration $p$\\
    \For{each patient $i \in \mathcal{B}_t$ }{
    $\Delta_t^i \gets \textbf{PatientUpdate}(i,\boldsymbol{\theta}_{t})$\\
    }
$\Delta_t \gets \dfrac{1}{|\mathcal{B}_t|}(\sum_i \Delta_t^i)$\\
$\Omega_{\sigma} = \{\sigma = z C_u / |\mathcal{B}_t| : \textbf{for}~z~\textbf{in}~\Omega_{z}\}$\\
$\Omega_{\Delta} = \{ \widetilde{\Delta} = \Delta_t + \mathcal{N}(0, (\sigma^2\mathbf{I})):\textbf{for}~\sigma~\textbf{in}~\Omega_{\sigma}\}$\\
$\widetilde{\Delta}_t \gets \textbf{NoisyUpdateSelect}(\Omega_{\Delta}, \epsilon', \mathcal{B}_t, \boldsymbol{\theta}_t, \mathcal{L})$\\
$\boldsymbol{\theta}_{t+1}=\boldsymbol{\theta}_t+ \widetilde{\Delta}_t$\\
}
\SetKwFunction{FMain}{PatientUpdate}
\SetKwProg{Fn}{Function}{:}{}
\Fn{\FMain{$i$, $\boldsymbol{\theta_t}$}}{
$\boldsymbol{\theta} \gets \boldsymbol{\theta}_t$\\
\For{batch image samples $b$ from Patient $i$}{
$\boldsymbol{\theta} \gets \boldsymbol{\theta} - \gamma \nabla \mathcal{L}(b;\boldsymbol{\theta})$
}
$\Delta^i = \boldsymbol{\theta} - \boldsymbol{\theta}_t$\\
$\Delta^i = ClipNorm(\Delta^i, {\rm C_u})$\\
return $\Delta^i$\\
}
\end{algorithm}
\begin{algorithm}
\SetKwFunction{FMain}{NoisyUpdateSelect}
\SetKwProg{Fn}{Function}{:}{}
\Fn{\FMain{$\Omega$, $\epsilon$, $\mathcal{B}$, $\boldsymbol{\theta}$, $\mathcal{L}$}}{
$\Omega_{u}$ = \{ u = $-Clip(\mathcal{L}$($\mathcal{B}$;$\boldsymbol{\theta} + \Delta$), ${\rm C_o}$) : \textbf{for} 
$\Delta$ \textbf{in} $\Omega$ \}\\
Select $\Delta$ with probability $\dfrac{exp(\dfrac{\epsilon u}{2{\rm C_o}})}{\sum_{u\in \Omega_u} exp(\dfrac{\epsilon u}{2{\rm C_o}})}$\\
return $\Delta$\\
}
\caption{NoisyUpdateSelect}
\label{algorithm:dpSelect}
\end{algorithm}

For \emph{update computation}, at the beginning of the $t$-th step of P3SGD, we randomly sample a patient batch $\mathcal{B}_t$ from the database $\mathcal{D}$ with a sampling ratio $p$. Here, the notation $\mathcal{B}_t$ is different from the one in 
Equation~\ref{eq:sgd}, where the $\mathcal{B}_t$ is sampled from individual images instead of patients.

Then, for each patient $i$ in the sampled batch, we perform a back propagation to calculate gradients of the parameters via images of the patient $i$. After that, we locally update the model using the computed gradients. After we traverse all images of this patient, we
can obtain the model update with respect to patient $i$.
This procedure can be interpreted as performing SGD on the local data
from patient $i$. 

In the next step, we average updates of all patients in $\mathcal{B}_t$ to obtain the final update at the $t$-th step.
Note that we need to control the sensitivity 
of the total update for further \emph{update sanitization}.
In practice, this is implemented by clipping the
$l2$ norm of the update, with respect to each individual patient
~(as shown in line~$26$ in Algorithm~\ref{algorithm:main}).
${\rm C_u}$ in Algorithm
~\ref{algorithm:main} denotes a predefined upper-bound. Thus, the sensitivity of the total update can be
bounded by $2{\rm C_u}$~(a proof can be found in supplementary materials).
The main idea of \emph{update computation} is implemented by
a function \textbf{PatientUpdate}, as shown in Algorithm~\ref{algorithm:main}.
 
To protect privacy, \emph{update sanitization} needs to be performed. Specifically, we use Gaussian 
mechanism~\cite{dwork_book} to inject well-calibrated Gaussian noise into the original update, which leads to a noisy update. 
The variance of injected Gaussian noise is jointly determined by the upper-bound ${\rm C_u}$ of the update's $l2$ norm and the noise scale $z$. 
In this paper, we use a common strategy to set ${\rm C_u}$ as a globally fixed
value similar to prior works~\cite{abadi2016deep,brendan2018learning}. 
Therefore, the choice of a noise scale factor $z$ is critical to train CNN model with high performance.
Previous works~\cite{abadi2016deep, brendan2018learning} usually use a fixed noise scale throughout the training phase.
However, the fixed noise scale factor may lead
to the departure of the noisy update from the descent direction or an ignorable
regularization effect, because the
magnitudes of the updates may vary at different iterative steps.
Thus, we argue that the strategy that uses a fixed noise scale may hinder the classification performance. 

In this paper, we present an elaborated strategy to adaptively select the noise
scale. This strategy is originated from
the exponential mechanism~\cite{dwork_book}, which is
a commonly used mechanism to build a differentially private version of the \textbf{Argmax} function. In this paper,
the \textbf{Argmax} function refers to select the argument which maximizes a specific objective function. In our task, we use the negative loss function as the objective function, and the argument is the noisy update built upon
different noise scales from the predefined set $\Omega_{z}$.
We implement this strategy as a function \textbf{NoisyUpdateSelect} depicted in Algorithm~\ref{algorithm:dpSelect}. 
The predefined set $\Omega_{z}$ 
contains $N_z$ noise scale factors. Increasing 
$N_z$ leads to more subtle control of the noisy update, which further boosts the performance. However, the increase of $N_z$ also results in an increase  of computational cost. Precisely,
one more noise scale will bring about one more forward computation on all
images in $\mathcal{B}_t$. 
In practice, we find that setting $N_z = 2$ suffices for our task. 
Note that setting $N_z = 1$ degenerates to the method used in~\cite{brendan2018learning, geyer2017differentially}.
In the experiments, we show this strategy is
crucial to boost the performance.

For \emph{privacy accumulation}, the composition theorem can be leveraged to compose the privacy cost at each iterative step. 
In this paper, we make use of the moments accountant~\cite{abadi2016deep}, which can obtain tighter bound than previous strong composition theorem~\cite{2010strong}. Specifically, the moments accountant is to track a
bound of the privacy loss random variable 
instead of a bound on the original privacy 
budget. Given a randomized algorithm $\mathcal{A}$, the privacy loss at output $o$
is defined as:
\begin{equation}
c(o;\mathcal{A}, \mathbf{aux}, \mathcal{D}^{'}, \mathcal{D}^{''}) \triangleq  \log \dfrac{\mathbf{Pr}[\mathcal{A}(\mathbf{aux}, \mathcal{D}^{'}) = o]}{\mathbf{Pr}[\mathcal{A}(\mathbf{aux}, \mathcal{D}^{''})=o]}
\end{equation}
Then, the privacy loss random variable $C(\mathcal{A}, \mathbf{aux}, \mathcal{D}^{'}, \mathcal{D}^{''})$ is defined
by evaluating the privacy loss at the outcome sampled from $\mathcal{A}(\mathcal{D}')$~\cite{pate}.
Here, $\mathcal{D}^{'}$ and $\mathcal{D}^{''}$ are adjacent.
$\mathbf{aux}$ denotes the auxiliary information. In our P3SGD algorithm, auxiliary information at step $t$ is the
weight parameters $\boldsymbol{\theta}_{t-1}$ at the step $t-1$. The algorithm 
$\mathcal{A}$ is also known as the adaptive mechanism in literature ~\cite{abadi2016deep}.
We can then define the moments accountant as follows:
\begin{equation}
    M_c(\lambda) \triangleq \max_{\mathbf{aux}, \mathcal{D}', \mathcal{D}^{''}} M_c(\lambda; \mathbf{aux}, \mathcal{D}^{'}, \mathcal{D}^{''})
\end{equation}
where $M_c(\lambda; \mathbf{aux}, \mathcal{D}^{'}, \mathcal{D}^{''})$ is the moment generating function
of the privacy loss random variable, which is calculated as:
\begin{equation}
    M_c(\lambda; \mathbf{aux}, \mathcal{D}^{'}, \mathcal{D}^{''}) \triangleq~\log~\mathbb{E}[\exp(\lambda C(\mathcal{A}, \mathbf{aux}, \mathcal{D}^{'}, \mathcal{D}^{''}))]
\end{equation}

Then, we introduce the composability and the tail bound of moments accountant as:
\begin{myTheorem}
 \textbf{(Composability)} Suppose that a randomized algorithm $\mathcal{A}$ consists of a sequence
 of adaptive mechanisms $\mathcal{A}_1, \dots, \mathcal{A}_k$ where $\mathcal{A}_i:\prod_{j=1}^{i-1} \mathcal{R}_j \times \mathcal{D}^{'} \rightarrow \mathcal{R}_i$. The moments accountant of $\mathcal{A}_i$ is denoted
 as $M_c^i(\lambda)$. For any $\lambda$:
 \begin{equation}
     M_c(\lambda) \leq \sum_{i=1}^{k}M_c^i(\lambda)
 \end{equation}
 \label{theorem:compo}
\end{myTheorem}
\vspace{-0.4cm}
 \vspace{-0.2cm}
\begin{myTheorem}
 \textbf{(Tail bound)} For any $\epsilon \ge 0$, the algorithm $\mathcal{A}$ satisfies $(\epsilon, \delta)$-differential
 privacy for
 \begin{equation}
     \delta = \min_{\lambda}\exp(M_c(\lambda)-\lambda \epsilon)
 \end{equation}
 \label{theorem:tail}
\end{myTheorem}
\vspace{-0.8cm}
Theorem~\ref{theorem:tail} indicates that if the moments accountant of a randomized algorithm $\mathcal{A}$ is bounded, then $\mathcal{A}$ satisfies $(\epsilon, \delta)$-differential privacy. The bound of the moments accountant for our strategy implemented in Algorithm~\ref{algorithm:dpSelect} is guaranteed by the following
theorem:

\begin{myTheorem}
 Given $\lambda$, the moments accountant of Algorithm~\ref{algorithm:dpSelect} is bounded by 
 $q\cdot \dfrac{\lambda(\lambda + 1)\epsilon^2}{2}$.
\label{theroem:dpSelect}
\end{myTheorem}
The proof can be done using the privacy amplification~\cite{amplicication11} and the theorem in the prior literature~\cite{bun2016concentrated}. More details can be found in appendix of this work.

\noindent \textbf{Privacy guarantee:}~In this paper, \emph{privacy accumulation}
is to accumulate the moments accountant's bound at each step. 
Note that privacy accumulation needs to be performed at the noisy update selection (line $17$ in Algorithm~\ref{algorithm:main})
and the
model update via noisy update~(line $18$ in Algorithm~\ref{algorithm:main}).
For the \textbf{NoisyUpdateSelect} in line $17$, we can calculate
a bound via Theorem~\ref{theroem:dpSelect}. For the 
model update in line $18$, the bound is obtained based on the property of Gaussian Mechanism (Lemma 3 in appendix of \cite{abadi2016deep}). 
Once we bound the moments accountant
at each iterative step, we can compose these bounds using Theorem~\ref{theorem:compo}. At last, the total privacy cost  is obtained based on Theorem~\ref{theorem:tail}. It suffices to compute
the $M_c(\lambda)$ when $\lambda \leq 32$. In practice, we use a finite set $\{1,\cdots, 32\}$ following prior work~\cite{abadi2016deep}.

\section{Experimental Results}
\label{section:results}
\subsection{Experimental Settings}
\begin{table*}[t]
\centering
\vspace{1ex}
\scalebox{1.0}{
\begin{tabular}{|l|c|c| |c|c|c|| c|c|c|     |c|c|c|} \hline

\multirow{2}{*}{Model}&\multirow{2}{*}{Type}&\multirow{2}{*}{\# Params} & \multicolumn{3}{c||}{SGD} & \multicolumn{3}{c||}{SGD+Dropout}                                                               & \multicolumn{3}{c|}{P3SGD}                                           \\ \cline{4-12} 
                       & & &\multicolumn{1}{c|}{Training} & \multicolumn{1}{c|}{Testing}& Gap  & 
                       \multicolumn{1}{c|}{Training} & \multicolumn{1}{c|}{Testing} & Gap                            & \multicolumn{1}{c|}{Training} & \multicolumn{1}{c|}{Testing}       & Gap   \\ \hline 
AlexNet  &\textbf{T} & $60.9$ M   & $99.87$ & $91.58$ &$8.29$   & $98.97$                        & $\mathbf{93.13}$                      & $5.84$                           & $96.85$                         & $92.74$                           & $4.11$ \\ \hline
VGG-16  &\textbf{T} & $14.7$ M                               & $99.81$& $92.19$&$7.81$                        &  $99.28$ & $\mathbf{94.32}$                          & $4.96$                      & $96.23$ & $93.87$  & $2.36$ \\ \hline
ResNet-18  & \textbf{M}& $11.2$ M                    & $99.85$ & $92.25$& $7.60$                       & $99.63$                          & $92.12$                      & $7.51$                           & $95.70$                         & $\mathbf{95.23}$                           & 0.47 \\ \hline
ResNet-34   &\textbf{M}& $21.3$ M                         & $99.23$ &$93.19$ & $6.04$                        & $99.16$                          & $93.22$ & $5.94$                          & $95.80$& $\mathbf{95.34}$         & $0.46$ \\ \hline
MobileNet    & \textbf{M} &$3.2$ M                          & $98.73$ & $92.01$& $6.72$                      & 
$98.65$ & $91.61$ 
& $7.04$& $94.79$   & $\mathbf{94.13}$            & $0.66$ \\ \hline
MobileNet v2 & \textbf{M} & $2.3$ M                           & $98.52$& $93.24$& $5.28$ & $98.37$                          & $93.28$  & $5.09$                          & $95.32$       & $\mathbf{94.86}$  & $0.46$\\ \hline
\end{tabular}
}
\caption{Training and testing accuracies (\%) of various network architectures trained with different strategies. The gap between training and testing accuracies is used for measuring the overfitting of the CNN models.The type \textbf{T}/\textbf{M} denotes traditional/modern CNNs.}
\label{table:all_results}
\end{table*}

In this section, we verify the effectiveness of P3SGD on a
real-world clinical dataset. This dataset is collected by the doctors in our team.
The dataset consists of $1216$ patients and each patient contains around $50$ image patches.
The task we consider in this paper is glomerulus classification, which aims to classify whether
an image patch contains a glomerulus or not. This task has also been studied in a recent work
~\cite{Gallego2018Glomerulus}.
We ask the doctors to manually label the image patches. 
For a fair comparison, we set the weight decay to 1e-4 and use data augmentation in all experiments.
Specifically, we perform data augmentation by (1) randomly flipping input
images vertically and/or horizontally, and (2) performing
random color jittering, including changing the brightness
and saturation of input images. All input images are
resized into $224\times224$, and pixel intensity values are 
normalized into $[0,1]$. All $1216$ patients in the dataset are randomly split into a training dataset~($1000$ patients) and a testing dataset ($216$ patients). 

\subsection{Classification Evaluation}
To validate the superiority of P3SGD in reducing overfitting, we compare it with the standard SGD (without Dropout). We also provide comparisons 
with the strategy that combines the standard SGD with Dropout. 
As a result, there are three training strategies:
SGD, SGD+Dropout, and P3SGD.

We first evaluate our method on the ResNet-18 architecture~\cite{he2016deep}.
For the standard SGD with Dropout, we insert Dropout between convolutional layers and set the drop ratio to $0.3$ following~\cite{wider-resnet}.
To provide a reasonable weight initialization, we firstly pre-train the CNN model
on a publicly available pathological image dataset\footnote{http://www.andrewjanowczyk.com/use-case-4-lymphocyte-detection/}. The pre-training does not take an extra privacy cost, since we do not interact with the original training dataset in this stage. The pre-training can also help us to determine the hyper-parameters in Algorithm~\ref{algorithm:main}. For P3SGD, we set the total updating rounds $T$ to $100$ and set the noise scale $\epsilon^2$ to $0.1$ for selecting noisy update.
The sampling ratio $p$ is set to $0.1$ and $\Omega_{z}$ is set to be $\{3.0, 1.0\}$. The
${\rm C_u}$ and ${\rm C_o}$ are set to $5.0$ and $3.0$, respectively.
To facilitate the discussion, we denote {\tt SGD} and {\tt P3SGD}
as the models trained using SGD (without Dropout) and P3SGD~(we use the abbreviations
in the following discussions). 

\begin{figure}
    \centering
    \includegraphics[width = 7cm]{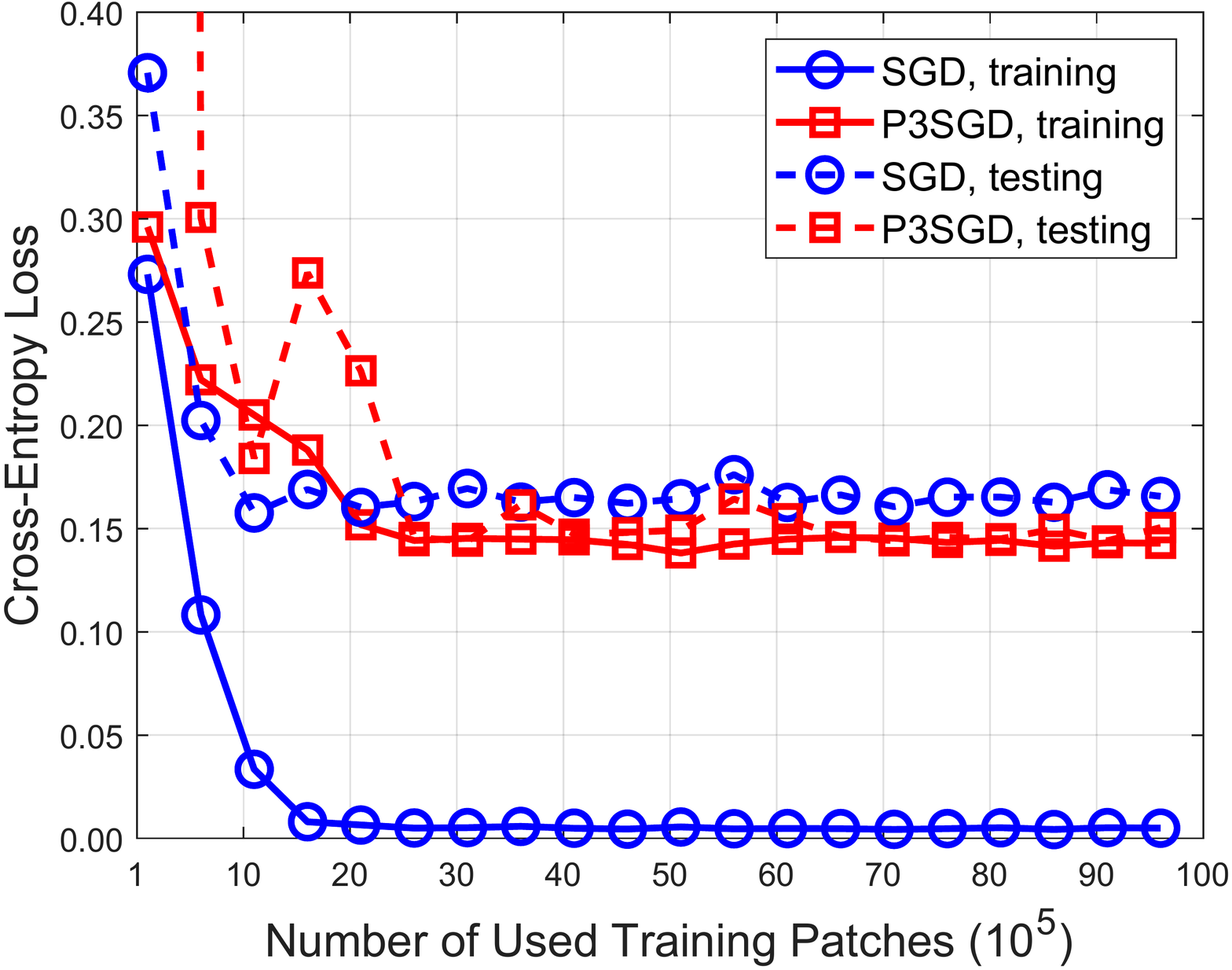}
    \caption{The training and testing loss curves of ResNet-18 with different training strategies. P3SGD significantly reduces the overfitting compared to SGD.}
    \label{fig:loss_compare}
    \vspace{-0.4cm}
\end{figure}

From the results of ResNet-18 (Table~\ref{table:all_results}), we observe that {\tt SGD} obviously overfits (it even reaches nearly $100\%$ training accuracy). In contrast, {\tt P3SGD} drastically decreases the gap between training and testing accuracies and
improves the testing accuracy. 
In particular, {\tt P3SGD} outperforms {\tt SGD}  by $2.98\%$ in the testing
accuracy (a $38.5\%$ relative drop in classification error), while the gap is decreased
from $7.60\%$ to $0.47\%$, which shows a $93.8\%$ relative improvement. These results indicate that P3SGD significantly reduces the overfitting in the ResNet-18 model compared with the standard SGD. 
We also plot the loss curves of ResNet-18 in Figure~\ref{fig:loss_compare}, which further demonstrates the regularization effect of P3SGD. 
Besides, \textbf{ there is no significant performance improvement when we apply Dropout on ResNet-18}. Dropout even leads to a slight decrease~(from $92.25\%$ to $92.12\%$) in testing accuracy. 
We will discuss this phenomenon in details in Section~\ref{sec:discussion}.

Besides the ResNet-18, we also conduct extensive experiments on other popular CNN architectures. 
In general, we mainly test on two types of CNN models, namely, traditional CNNs and modern CNNs~\footnote{The modern CNN consists
of convolutional layers except the final prediction layer, which comprises of a global average pooling and a fully-connected layer.}  (denoted by \textbf{T} and \textbf{M} in Table~\ref{table:all_results}). 
Specifically, six architectures are included: AlexNet~\cite{2012alexnet}, VGG-16~\cite{2014vgg}, ResNet-18~\cite{he2016deep}, ResNet-34~\cite{he2016deep}, MobileNet~\cite{2017mobilenets}, and MobileNet v2~\cite{2018mobilev2}. For traditional CNNs~(\emph{e.g.},
AlexNet), we insert Dropout between fully connected~(FC) layers and set the drop ratio to $0.5$ following~\cite{dropout14}. The results are summarized in Table~\ref{table:all_results}. On the one hand, our method consistently boosts the testing accuracy over the standard SGD (without Dropout) on various CNN architectures. The ResNet-34 trained with P3SGD achieves the highest testing accuracy at $95.34\%$ among all network architectures and training strategies. In particular, \textbf{P3SGD outperforms
Dropout technique on all modern CNNs}, \emph{e.g.}, 
the testing accuracy gain is $2.12\%$ in the case of ResNet-34.
On the other hand, the training accuracy is suppressed when we use P3SGD to train the CNN model, which further leads to a decrease of the gap between training and testing accuracy.

Despite the superiority of our method, we observe that 
\textbf{Dropout is usually more effective than P3SGD on the traditional CNNs}, \emph{e.g.}, it obtains a slight accuracy gain of $0.45\%$ in the case of VGG-16 compared to P3SGD. We provide
some interpretations in the discussion part. 
We also notice that, under the standard SGD (without Dropout) training strategy, the modern CNNs have less overfitting (measured as the gap between training and testing accuracies) than traditional CNNs. This may be caused by the regularization effect brought by the Batch Normalization~\cite{batchnorm} which exists in the modern CNNs.

\subsection{Privacy Cost Analysis}

Another advantage of P3SGD is to provide patient-level
privacy within differential privacy. The differentially private degree is measured by $(\epsilon, \delta)$ (\emph{i.e}, privacy cost) in Equation~
\ref{eq:dp}. In this part, we calculate the total spend of privacy cost via the moments accountant theorem. 
The target  $\delta$ is fixed to $\dfrac{1}{|N_p|^{1.1}}$ ($N_p$
is the number of patients in the training set), which is suggested 
by the previous literature~\cite{dwork_dp_2006}. In our task, the $\delta$ is
around $5e-4$ ($N_p = 1000$).
To verify the effectiveness of our proposed strategy for dynamically controlling the noisy update, we compare it with the strategy of fixed
noise scale (marked by \ding{55} in Table~\ref{tab:privacy_cost}) which is adopted by the state-of-the-art works~\cite{geyer2017differentially,brendan2018learning}. For simplicity, we use {\tt adaptive} and {\tt fixed} to denote
these two strategies. 
All the experiments are performed on ResNet-18. 

We test on various noise scale sets $\Omega_z$ to show how
the noise scale affects the performance. 
We find that the noise scale greater than
$3.0$ leads to unstable training. In practice, we build $\Omega_z$ using the noise scale from $\{1.0, 2.0, 3.0\}$.
Overall, P3SGD with the {\tt adaptive} strategy ($\Omega_z=\{3.0, 1.0\}$) achieves the best testing accuracy of $95.23\%$ at a privacy cost of $6.97$.
For the {\tt fixed} strategy, a larger noise scale leads to a lower privacy
cost, however, it may cause the noisy update deviating from the decent direction and further hinders the testing accuracy. For example,
setting $\Omega_z$ to $\{3.0\}$ leads to the lowest privacy cost
of $4.70$ and the worst accuracy of $92.15\%$, 
while setting $\Omega_z=\{1.0\}$ achieves a better accuracy of $94.38\%$ but a much higher privacy cost of $8.48$. 
The {\tt adaptive} strategy provides a reasonable solution for this dilemma of the {\tt fixed} strategy. 

In general, the {\tt adaptive} strategy
leads to a better trade off between
the privacy cost and the testing accuracy.
Specifically, extending the fixed scale $\{z_1\}$ or $\{z_2\}$ to $\{z_1, z_2\}$ achieves the testing accuracy higher than or approaching to the 
best testing accuracy among the corresponding {\tt fixed} strategies, 
while with a reasonable privacy cost. 
For instance, the {\tt adaptive} strategy with $\{3.0, 1.0\}$ achieves the accuracy of $95.23\%$, which is higher than the {\tt fixed} strategy with either $\{3.0\}$ or $\{1.0\}$. Our strategy also outperforms a naive solution by setting the noise scale to the average of $1.0$ and $3.0$ (\emph{i.e.}, $\{2.0\}$). There is even an accuracy gain of $0.85\%$ by extending $\{1.0\}$ to $\{3.0, 1.0\}$. We infer this accuracy gain comes from the stronger regularization effect brought by the larger noise scale.
Meanwhile, the {\tt adaptive} strategy ($\{3.0, 1.0\}$) achieves a moderate privacy cost between the costs obtained by
the corresponding {\tt fixed} strategies (setting $\Omega_z$
to $\{1.0\}$ or $\{3.0\}$).

\begin{table}[t]
    \centering
    \begin{tabular}{|c|c||c||c|}
         \hline
         Adaptive& $\Omega_{z}$&Testing& $\epsilon$   \\
         \hline
         \text{\ding{51}} & $\{3.0, 1.0\}$& $\mathbf{95.23}$ & $6.97$\\
         \hline
         \text{\ding{51}} & $\{2.0, 1.0\}$& $94.31$ & $7.10$ \\
         \hline
         \text{\ding{51}} & $\{3.0, 2.0\}$& $93.57$ & $4.97$\\
         \hline
         \text{\ding{55}}  & $\{1.0\}$  &   $94.38$ & $8.48$ \\
         \hline
         \text{\ding{55}}  & $\{2.0\}$ & $93.24$ & $5.13$ \\
         \hline
         \text{\ding{55}}  & $\{3.0\}$ & $92.15$ & $4.70$   \\
         \hline
    \end{tabular}
    \caption{Testing accuracy~($\%$) and privacy cost of different settings on
    ResNet-18. The \emph{Adaptive} column indicates if an adaptive noise scale set (marked by a \text{\ding{51}}) or a fixed scale (marked by a \text{\ding{55}}) is used. Setting a fixed noise scale leads to the method in~\cite{geyer2017differentially, brendan2018learning}}
    \label{tab:privacy_cost}
\end{table}

To conclude, our proposed strategy can be seen as a simplified version of line search in numerical optimization~\cite{06optimization}, and provides a more careful way to control the magnitude of the added noise. The effectiveness of our strategy comes from the fine-grained way to control the noisy update.

\begin{figure*}
    \centering
    \includegraphics[width=14cm]{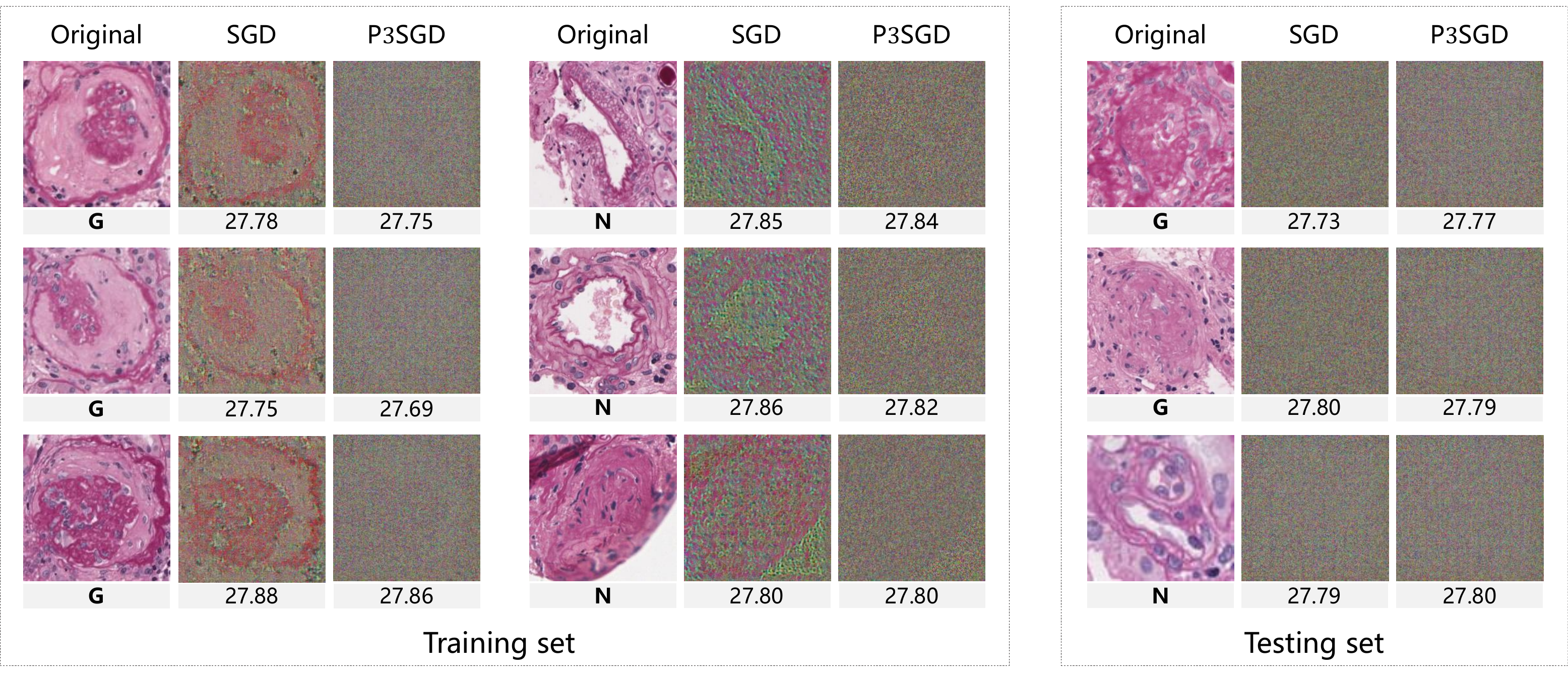}
 \caption{Visualization of the model-inversion attack. \textbf{G}/\textbf{N} below each original patch denotes if the patch contains a glomerulus or not. The number below each reconstructed image is the PSNR value. The reconstructed examples of the training dataset are demonstrated in the left part. For comparison, we also show some examples from the testing dataset in the right part. }
 \vspace{-0.4cm}
\label{fig:vis_all_inverse}
\end{figure*}
\begin{table}
    \centering
    \begin{tabular}{|c||c|c|c|}
         \hline
         Strategy&  Training & Testing& Gap\\
         \hline
         SGD+Dropout & $99.63$ & $92.12$ & $7.51$ \\
         \hline
         SGD+DropBlock& $98.85$ & $94.87$ &$3.68$  \\
         \hline
         P3SGD & $95.70$ & $\mathbf{95.23}$ & $0.47$ \\
         \hline
    \end{tabular}
    \caption{Training and testing accuracies~(\%) on ResNet-18 with different
    regularization strategies. }
    \label{tab:comparison_dropblock}
    \vspace{-0.4cm}
\end{table}

\subsection{Discussions}
\label{sec:discussion}
In this subsection, we first analyze the performance of different types of CNNs. We then compare P3SGD with the state-of-the-art regularization mechanism.  Finally, we show that the model trained with P3SGD is resistant to a model-inversion attack.

\nosection{Network Architecture}
As shown in Table~\ref{table:all_results}, 
Dropout and our method P3SGD demonstrate totally different effects on the two types of CNN architectures (traditional CNNs and modern CNNs). Specifically, our method outperforms 
Dropout on modern CNNs, instead, Dropout is more
effective on the traditional CNN architectures. This may be caused by following reasons: (1) Dropout is originally designed for the FC layers due
to its huge numbers of parameters~(\emph{e.g.}, around $90\%$ parameters of VGG-16 are from the FC layers). However, there is only one FC layer with a few parameters in modern CNN architectures. (2) The cooperation of Dropout and Batch Normalization can be problematic~\cite{batchnorm}. As we know, batch normalization layer widely exists in modern CNNs~(\emph{e.g}, ResNet~\cite{he2016deep}).
(3) Dropout discards features randomly, however,
the features extracted by convolutional layers are always spatially correlated,
which impedes the use of Dropout on convolutional layers. 
Some recent works propose to modify Dropout for
convolutional layers. We compare our method with a variant of Dropout in the next part. 

\nosection{Other Regularization Techniques}
From the previous discussion, 
some advanced forms of Dropout should be adopted in the modern
CNN. In this part, we compare our method with
a recent technique, named DropBlock~\cite{2018dropblock}, on ResNet-18. 
For a fair comparison with Dropout, we insert DropBlock between every two convolutional 
layers and set the drop ratio to $0.3$ following~\cite{2018dropblock}. The results are shown in Table~\ref{tab:comparison_dropblock}. DropBlock achieves a testing accuracy gain of $2.75\%$ against Dropout, while P3SGD
outperforms both Dropout and DropBlock. In contrast to P3SGD, DropBlock has no suppression effect on the training accuracy. We guess that the performance gain of DropBlock comes from the effect of the implicit model ensemble.
We further combine P3SGD
with DropBlock but do not obtain obvious accuracy boost. 




\nosection{Model-inversion Attack}
To demonstrate that {\tt P3SGD} is resistant to the model-inversion attack \cite{federated_learning,mahendran2015understanding}, we perform
an inversion attack on CNN models trained with different strategies. 
As a case study, 
we conduct experiments on the ResNet-18 and use the output features from the $3$-th residual block to reconstruct the input image (see details in the appendix).
Some visualizations are shown in Figure~\ref{fig:vis_all_inverse}. 
We can reconstruct the outline of the tissue in the input image using the features from the {\tt SGD}. In contrast, 
 we can not obtain any valuable information from {\tt P3SGD}
 (\emph{i.e.}, the model is oblivious to training samples).
It indicates that {\tt SGD} is more vulnerable than {\tt P3SGD}.
Quantitatively, we perform attack on all the training images and report the average
PSNR values as: $27.82$ for {\tt P3SGD} and $27.84$ for {\tt SGD}. We also conduct the same study on patches from the testing dataset 
and show some examples in the left part in
Figure~\ref{fig:vis_all_inverse}. The results show that it is hard to reconstruct the
input image for both {\tt SGD} and {\tt P3SGD}, since the testing examples are not touched by the model in the training phase. This provides some cues for the memorization ability of CNNs~\cite{Zhang2016a}.
\vspace{-0.2cm}
\section{Conclusion}
\label{section:conclusions}
\vspace{-0.2cm}
In this paper, we introduce a novel SGD schema, named P3SGD, to regularize the
training of deep CNNs while provide rigorous privacy protection
within differential privacy. P3SGD consistently outperforms
SGD on various CNN architectures. 
\textbf{The key technical innovation lies in the strategy that adaptively controls the noisy update}.
We conduct an analysis and show the effectiveness of this strategy.
We also perform a model-inversion attack and show that the model trained
with P3SGD is resistant to such an attack. 

This research paves a new way to regularize deep CNNs on pathological image analysis with an extra advantage of appealing patient-level privacy protection. Applying this method to other types of medical
image analysis tasks is promising and implies a wide range of clinical applications.


{\bf Acknowledgment}
Bingzhe Wu and Guangyu Sun are supported by National Natural Science Foundation of China (No.61572045).

{\small
\bibliographystyle{ieee}
\bibliography{egbib}
}

\end{document}